\documentclass[10pt,journal, twoside, web]{ieeecolor}
\usepackage{multirow}
\usepackage{amsmath,amsfonts}
\usepackage{algorithmic}
\usepackage{algorithm}
\usepackage{array}
\usepackage[caption=false,font=normalsize,labelfont=sf,textfont=sf]{subfig}
\usepackage{textcomp}
\usepackage{stfloats}
\usepackage{url}
\usepackage{verbatim}
\usepackage{graphicx}
\usepackage{cite}
\usepackage[symbol]{footmisc}
\usepackage{kotex}
\usepackage[flushleft]{threeparttable}
\let\labelindent\relax
\usepackage{enumitem}

\newcommand{\yh}[1]{{\textcolor{black}{#1}}}

\newcommand{\add}[1]{{\textcolor{black}{#1}}}
\newcommand{\rev}[1]{{\textcolor{black}{#1}}}

\makeatletter
\let\NAT@parse\undefined
\makeatother
\usepackage[numbers,compress]{natbib}
\bibpunct{[}{]}{,}{n}{}{;}
\def\BibTeX{{\rm B\kern-.05em{\sc i\kern-.025em b}\kern-.08em
    T\kern-.1667em\lower.7ex\hbox{E}\kern-.125emX}}

\begin{document}

\title{AI-driven View Guidance System in Intra-cardiac Echocardiography Imaging}

\author{{Jaeyoung Huh}, {Paul Klein}, {Gareth Funka-Lea}, {Puneet Sharma}, {Ankur Kapoor}, {Young-Ho Kim}
\thanks{J. Huh, P. Klein, G. Funka-Lea, P. Sharma, A. Kapoor, Y.-H. Kim is Digital Technology \& Innovation, Siemens Healthineers, Princeton, NJ, USA.}
\thanks{Y.-H. Kim is a corresponding author. \\{\tt\footnotesize (e-mail: young-ho.kim@siemens-healthineers.com)}}
}

\maketitle
\begin{abstract}
Intra-cardiac echocardiography (ICE) is a crucial imaging modality used in electrophysiology (EP) and structural heart disease (SHD) interventions, providing real-time, high-resolution views from within the heart. Despite its advantages, effective manipulation of the ICE catheter requires significant expertise, which can lead to inconsistent outcomes, especially among less experienced operators. To address this challenge, \yh{we propose an AI-driven view guidance system that operates in a continuous closed-loop with human-in-the-loop feedback, designed to assist users in navigating ICE imaging without requiring specialized knowledge.}
Specifically, our method models the relative position and orientation vectors between arbitrary views and clinically defined ICE views in a spatial coordinate system. It guides users on how to manipulate the ICE catheter to transition from the current view to the desired view over time. By operating in a closed-loop configuration, the system continuously predicts and updates the necessary catheter manipulations, ensuring seamless integration into existing clinical workflows. 
The effectiveness of the proposed system is demonstrated through a simulation-based performance evaluation using real clinical data, achieving an 89\% success rate with 6,532 test cases. \yh{Additionally, a semi-simulation experiment with human-in-the-loop testing validated the feasibility of continuous yet discrete guidance. These results underscore the potential of the proposed method to enhance the accuracy and efficiency of ICE imaging procedures}

\end{abstract}


\section{Introduction}

\IEEEPARstart{I}{ntra}-cardiac echocardiography (ICE) is a sophisticated imaging modality that provides real-time, high-resolution views from within the heart, making it an invaluable tool in both electrophysiology (EP) and structural heart disease (SHD) interventions. 
ICE is commonly used in procedures such as atrial fibrillation ablation, transcatheter valve repairs, and septal defect closures. The key advantages of ICE include its ability to provide clear and detailed near-field images of cardiac structures. It can be performed under conscious sedation, manipulated easily, and interfaced with other interventional equipment. Additionally, ICE has been shown to reduce procedural and fluoroscopy times, decrease overall radiation exposure to both the patient and physician, and shorten hospital stays. As a result, ICE has become widely used in interventional procedures for the management of complex cardiac conditions\,\cite{calo2002ep,saliba2008abl,basman2017shd,goya20ice}.

The primary use cases of ICE imaging involve visualizing target anatomy, detecting and tracking therapeutic devices, and validating treatments in real-time. For ICE imaging to be effective, the ICE catheter must be accurately positioned at the target location. Achieving this requires physicians to have significant expertise in interpreting anatomical views via ICE images and skillfully maneuvering the ICE catheter using two knobs (anterior-posterior, right-left). They must also rotate and translate the catheter body itself to reach the target anatomical views, while primarily focusing on therapeutic device manipulation\,\cite{tan2019chd,bartel2013device}. This complexity can be particularly challenging for those who are less experienced. As a result, the procedure is largely dependent on the skill of the operator, which might lead to inconsistent outcomes and extended procedure times. It emphasizes the importance of experience in achieving optimal results.

\begin{figure}[t!]
  \center
	\includegraphics[width=0.5\textwidth]{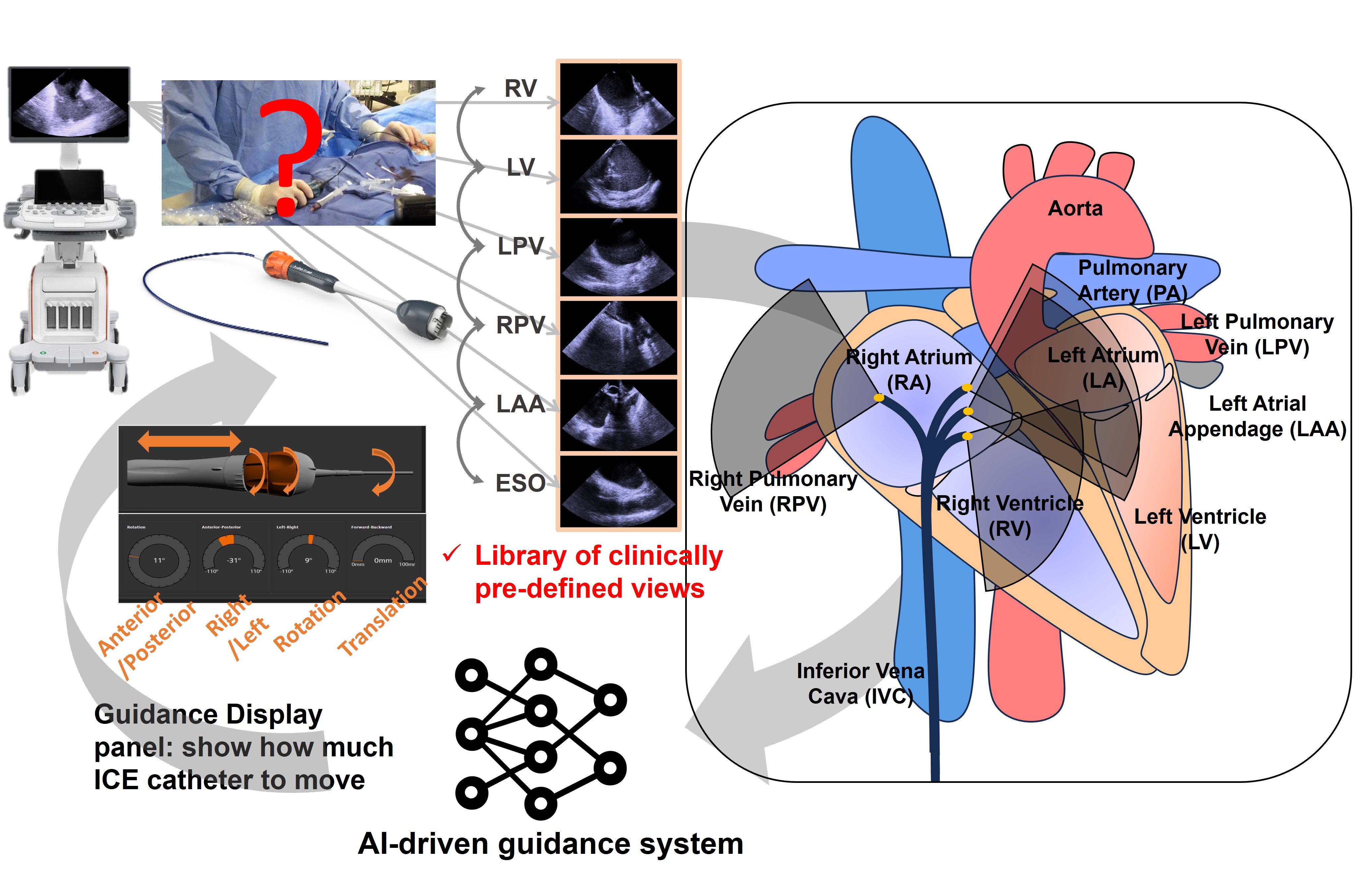}
	\caption{Overview of the proposed view guidance system. When users want to navigate to clinically predefined views during procedures, our proposed system provides continuous guidance on how to manipulate the ICE catheter via an interactive device ({\em e.g., } a touchpad for view selection and a feedback monitor) until the target view is achieved. (ACUSON Origin Cardiovascular Ultrasound System and AcuNav Lumos ICE catheter, Source: Siemens Healthineers)}
	\label{fig:whole}
\end{figure}

In this research, we propose an AI-driven view guidance system for continuous closed-loop guidance with human-in-the-loop feedback, designed to assist users in manipulating ICE imaging. The system aims to empower even novice users to effectively operate ICE imaging without requiring specialized expertise. Figure\,\ref{fig:whole} presents an overview of the proposed view guidance system. Specifically, our proposed method models the relative position and orientation vectors between arbitrary views and clinically defined common ICE views\,\cite{jnj2019iceview,enriquez2018use,alkhouli2018intracardiac} - such as the {Right Ventricle (RV), Left Ventricle (LV), Left Atrial Appendage (LAA), Left Pulmonary Vein (LPV), Right Pulmonary Vein (RPV), or Esophagus (ESO) views}-within a unified spatial coordinate system.
\rev{The proposed model can be utilized in a closed-loop configuration with a human-in-the-loop, predicting the necessary manipulations to the ICE catheter's state between the current view and a clinically defined target view (i.e., the goal view). The user is guided step-by-step to perform the required manipulations on the ICE catheter, with continuous instructions provided until the desired view is successfully achieved.}

\section{Related works and Background}
\subsection{US view navigation}
To improve clinical workflow, various research efforts have focused on developing guidance systems for tasks such as view finding and clinical decision-making. In particular, Artificial Intelligence (AI)-based  methods have been widely utilized to guide and support these workflows. For instance, \citet{narang2021utility} proposed novel software that provides real-time prescriptive guidance to novice operators for obtaining the Trans-Thoracic Echocardiographic (TTE) images. Similarly, \citet{sabo2023real} proposed a real-time guidance system to improve the echo-cardiographic acquisition process. \citet{pasdeloup2023real} developed a method to help users navigate toward preferred standard views during scanning. Additionally, \citet{amadoul24simulation} demonstrated a synthetic image generation pipeline for navigation in an echocardiography view classification experiment. Building on this, \citet{amadou2024goal} proposed a framework using reinforcement learning to guide novice sonographers in navigating to standard interventional views during Transesophageal echocardiography (TEE).

While many pioneering works have significantly improved clinical workflows, the aforementioned studies primarily focus on acquiring echocardiograms externally or rely on simulation data. In contrast, our research is the first to specifically address a view guidance system for the ICE catheter manipulation based on real clinical data. This approach requires real-time understanding of internal cardiac structures, providing continuous guidance through the interactive mapping of images to spatial and catheter joint spaces. Furthermore, unlike simulation-based methods, which may not fully capture the complexities of real clinical environments, our method is designed for seamless integration into actual clinical workflows, effectively guiding the catheter within the heart's intricate anatomy.

Although some robotic systems\,\cite{zhongyu21ice,kim2022automated,loschak2016algorithms} exist to assist with catheter control, they are often expensive and complex to integrate into the clinical workflow. These systems typically focus on reproducing previously saved views by manual view survey\,\cite{zhongyu21ice,kim2022automated} or rely on positional sensors within a spatial coordinate system\,\cite{loschak2016algorithms}. In contrast, our approach relies solely on ICE-imaging-based state estimation to guide the user in manipulating the ICE catheter within the control space, from the current view to the target view. This eliminates the need for positional sensors and avoids disruption to the existing clinical workflow.

\subsection{{From Transformer to the Mamba and its medical applications}}


To address the relationship between non-linear behaviors in images and associated multi-modal states, many recent studies have turned to AI, particularly leveraging a Transformer architecture\,\cite{ashish17attention}. The transformer has been widely adopted in vision tasks due to its powerful ability to understand context across entire images through its self-attention mechanism, as well as its scalability to large models and robustness. As a result, many tasks that traditionally relied on convolutional neural network (CNN) structures are now being addressed with transformer-based architectures\,\cite{dosovitskiy2020image,carion2020end,liu2021swin,touvron2021training,kim23ai}.

Recently, a new alternative to the transformer architecture, called Mamba\,\cite{gu2023mamba}, has emerged. Unlike the transformer, which relies on self-attention mechanisms with quadratic computational complexity for long sequences, the State Space Model (SSM) maintains linear computational complexity. The Mamba is based on the SSM, which can be represented as follows:
\begin{equation}
    \begin{split}
        h[k]& = \hat{A} h[k-1] + \hat{B} x[k]\\
        y[k]& = C h[k],
    \end{split}
\end{equation}
where the $h[k]$ represents the state at the $k$-th step, and the matrix $C$ is the projection matrix that projects the current state to the output. The matrices $\hat{A}$ and $\hat{B}$ are the discretized versions of the system matrix and the projection matrix, respectively. They can be represented as follows:
\begin{equation}
    \begin{split}
        \hat{A}&=\exp(\Delta A) \\ 
        \hat{B}&=(\Delta A)^{-1}(\exp(\Delta A)-I) \cdot \Delta B,
    \end{split}
\end{equation}
where the $\Delta$ denotes the time-scale parameter.

Mamba implements the Selective Scan Mechanism with SSM (S6), filtering out irrelevant information to effectively handle long-range sequences and induction heads. To achieve this, it allows the matrices $B$, $C$, and $\Delta$ to depend on the input sequence. Additionally, its hardware-aware implementation makes Mamba computationally efficient model, leading to significant performance gains and improved resource efficiency.

Due to its outstanding performance, many researchers have started replacing the transformer architecture with the Mamba structure. Recently, numerous efforts have been made to adapt the Mamba structure to the medical domain for tasks such as segmentation, classification, and detection. For instance, the authors in \cite{architvim,ma2024u,xing2024segmamba,ruan2024vm,wang2024weak} proposed Mamba-based methods for medical image segmentation. \citet{yue2024medmamba} introduced a medical image classification method that combines CNN with the Mamba structure. In their approach, local features are extracted using CNN, while global features are obtained from the Mamba structure, which are then combined to classify various medical images. In \cite{guo2024mambamorph}, the author presented MambaMorph, a Mamba-based multi-modality deformable registration framework. Additionally, \citet{schiff2024caduceus} utilized bi-directional Mamba blocks to model genomic sequences. 

Our goal is to find and match the global features of ICE images to coordinates within a spatial coordinate system, which requires leveraging the entire content of the image. In this context, we believed that the Mamba structure offers superior performance compared to transformers, and we have integrated it into our research.

\rev{Building on these pioneering works, we propose a novel ICE view guidance system with the following key contributions:}
\begin{enumerate}[itemsep=0mm]
    \item \add{We introduce a new ICE view guidance system that enables precise navigation from arbitrary views to clinically defined target views.}
    \item \add{Our approach leverages a variation of the Mamba regression model, enabling robust and accurate guidance.} 
    \item \add{We validated the feasibility of the proposed system through comprehensive simulation and semi-simulation experiments, effectively demonstrating its real-world applicability.}
\end{enumerate}

\section{Materials and Methods}

\subsection{Problem Setup and Notation}

The ICE catheter is manually inserted into the femoral vein at the groin through an introducer sheath. A basic ICE study typically begins with the catheter positioned in the mid-right atrium (RA), achieving the {\it home view}, with the catheter in a neutral position. It provides imaging of the RA, tricuspid valve, and right ventricle\,\citep{briceno20ice}. Clinicians often rely on this {\it home view} as a reference that can be easily restored if difficulties arise during ICE imaging. We assume this {\it home view} as the reference coordinate system for our proposed method.

Since most ICE catheter procedures are conducted in the RA, our guidance system assumes that there are no obstacles in the RA, and therefore, obstacle avoidance is not addressed in this article.
\yh{Based on the physiological properties of the cardiac cycle, we assume that ICE imaging is captured during the diastolic phase, specifically at the end of the T-wave on the ECG. This timing corresponds to when the heart is fully relaxed, the ventricles are filled with blood, and the Left Atrium (LA) reaches its maximum extent. While this assumption aligns with common settings in systems like CARTO\,\cite{jnj2023carto} (Detailed in Section\,\ref{sec:dataset}), it provides a stable and consistent reference point for imaging, minimizing motion artifacts\,\cite{yeh08qrs}.}

Finally, while we use the tip's position and orientation information for modeling, it is important to note that our actual guidance system does not require position/orientation sensors attached to the ICE catheter. This allows our system to operate as a purely ICE-image-based guidance system, eliminating the need for additional sensor attachments on the catheter.

We define the ICE imaging state {\bf S} = {($x, y, z, \delta x, \delta y, \delta z$)} as the position and orientation in $\mathbb{SE}(3)$. The ICE catheter has four degrees of freedom: two DOFs for steering the catheter tip in two planes (anterior-posterior knob angle $\theta_1$ and right-left knob angle $\theta_2$) using two knobs on the catheter handle, bulk rotation $\theta_3$, and translation $d_4$ along the major axis of the catheter. Thus, the ICE joint state is {\bf J} = $(\theta_1, \theta_2, \theta_3, d_4)$. The ICE imaging, {\bf I}, is defined as the acquired images, which can be either 2D or 3D. 

Since the {\it home view} is a neutral position and one of the predefined clinical views, we define a transition function from {\bf S} to {\bf J} using inverse kinematics ($\mathcal{IK}$)\,\cite{kim2020automatic}, with the home view (i.e., ${\bf S}_{home}$) serving as the reference coordinate system:

\begin{equation}~
{\bf J}_{home}^i= \mathcal{IK}({\bf S}_{home}^{i}). \label{def:IK} 
\end{equation}
{where $i$ represents the index of the possible ICE views.}

Then, we can define $\Delta {\bf J}_{(i,j)}$ as follows:
\begin{equation}~
\Delta {\bf J}_{(i,j)}= \mathcal{IK}({\bf S}_{home}^{i}) - \mathcal{IK}({\bf S}_{home}^{j}), \label{def:delta} 
\end{equation}
where $j$ represents the index of the possible ICE views.

{\noindent \bf Problem Definition:} Given the current imaging ${\bf I}{curr}$ and a goal index $g$ from a library of views, the objective is to estimate the relative ICE joint state, $\Delta {\bf J}{(curr,g)}$, over time:
\begin{eqnarray}~
S_{home}^{g} = \mathcal{M}({\bf I}_{home} , g), \label{eq:model1}\\
S_{curr}^{home} = \mathcal{M}({\bf I}_{curr}, home), \label{eq:model2}\\
\Delta {\bf J}_{(curr,g)}= \mathcal{IK}({\bf S}_{home}^{g}) - \mathcal{IK}(({\bf S}_{curr}^{home})^{-1}), \label{eq:model3}
\end{eqnarray}
where $\mathcal{M}$ is our proposed model, which will be detailed in the following sections (Sections\,\ref{sec:dataset} to \ref{sec:implementation}).

 \begin{figure*}[t!]
  \center
	\includegraphics[width=0.93\textwidth]{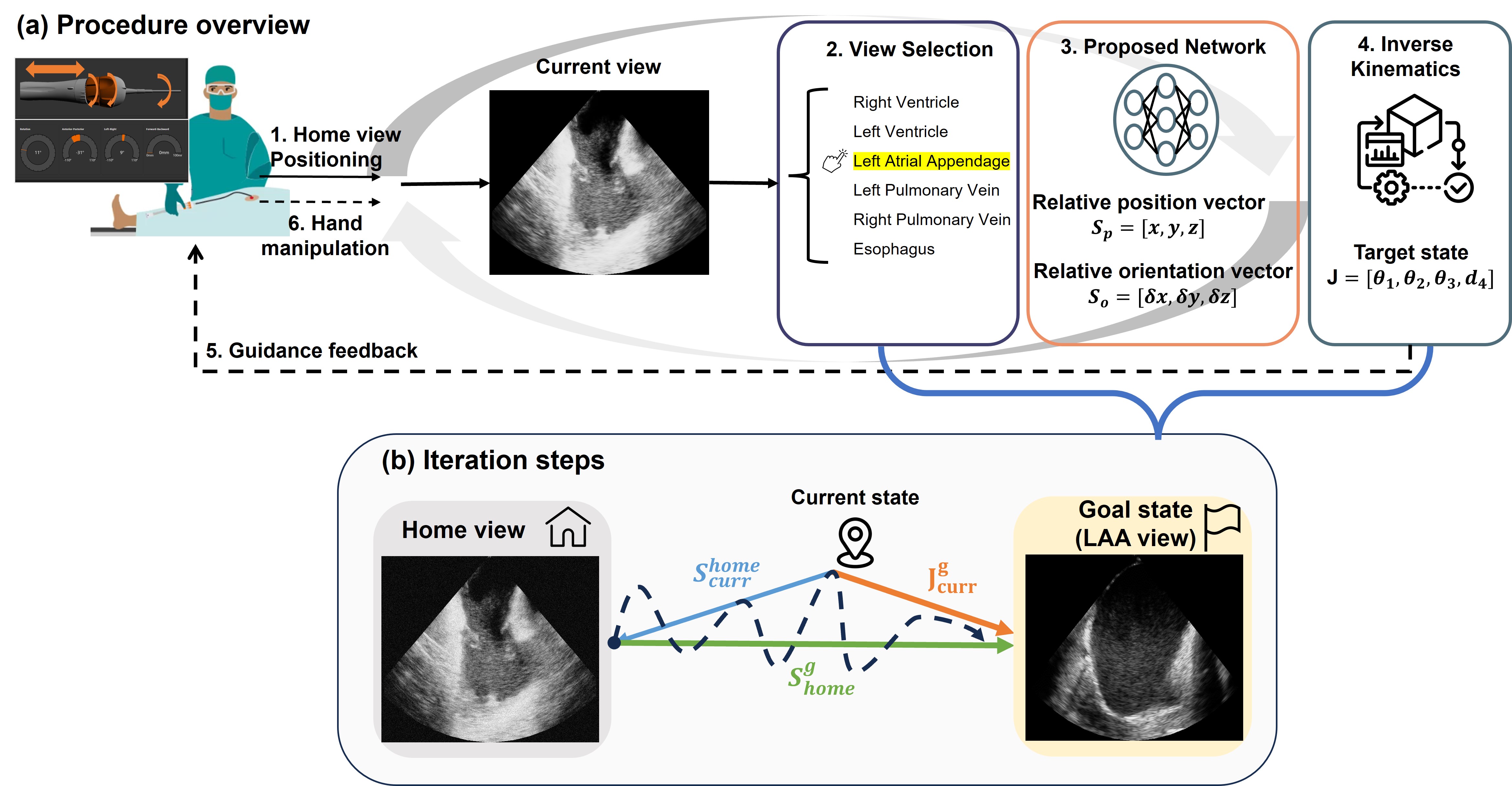}
	\vspace*{-0.15cm}
	\caption{Process of the proposed method. (a)Procedure Overview: The clinician first positions the ICE catheter at the home view and then selects the desired viewpoint. The proposed network provides the relative state, which is then transformed into robotic coordinates using inverse kinematics (IK). Guided by this information, the clinician manipulates the ICE catheter, and the procedure is performed iteratively. (b) Iteration step: During the procedure, the user can sequentially move toward the guided target state. The proposed method continuously updates the current state based on the home view state, enabling progress tracking from the starting point to the target point. Each step can be repeated as needed until the target view is achieved.}
 	\vspace*{-0.35cm}
	\label{fig:framework}
\end{figure*}

\begin{figure*}[t!]
  \center
	\includegraphics[width=\textwidth]{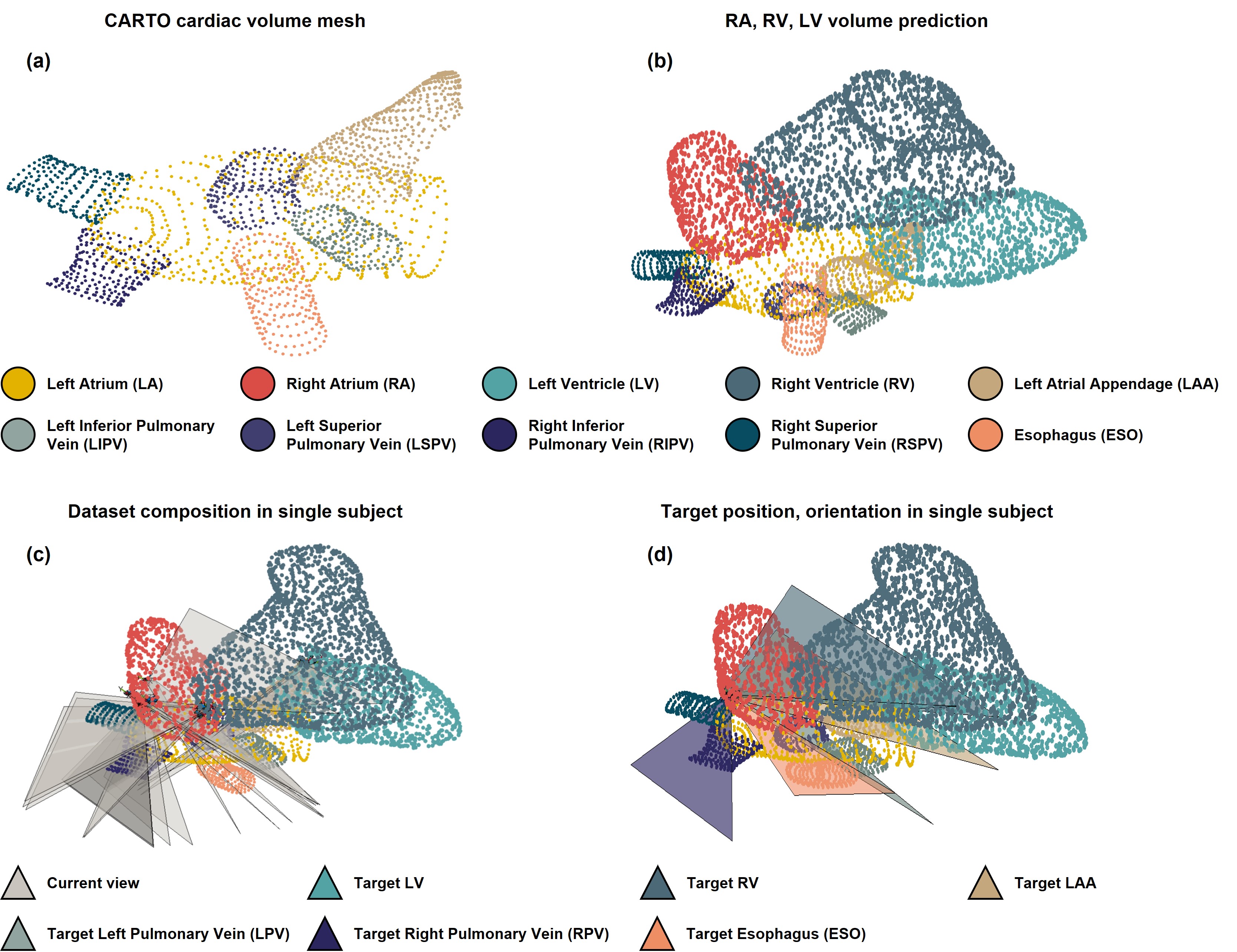}
	\caption{\add{(a) The reconstructed cardiac anatomical volume mesh of a single subject, generated using our volume contouring algorithm \cite{liao2018more}. (b) The reconstructed volume mesh for missing structures such as the RA, RV, and LV, where the mesh centers were predicted based on a small existing dataset. (c) The composition of the CARTO dataset for a single subject, which includes multiple image views visualizing different structures in various positions, along with corresponding position/orientation data and the volume mesh. (d) Our target dataset consists of six target states, each corresponding to one of six clinically defined views for a single subject. While each target state shares the same position, the orientations differ, with the fan direction aligned to the center of the target volume.}}
 	\vspace*{-0.35cm}
	\label{fig:dataset}
\end{figure*}

\subsection{Procedure Overview}
Figure\,\ref{fig:framework}~(a) provides an overview of the proposed ICE view guidance system. The procedure is divided into several key steps, each of which is detailed below:
\begin{enumerate}
\item {\bf Home view positioning:} The procedure begins by positioning the ICE catheter in the mid-RA to achieve the {\it home view}. Next, the ICE catheter is placed in a neutral position, which serves as the reference coordinate system.
\item {\bf The desired view selection:} The clinician selects the goal view (indexed as $g$) from a predefined library of clinical views. The selected view index is then fed into our model $\mathcal{M}$ to estimate the current imaging state in spatial space $S_{home}^{g}$ (Equation\,\eqref{eq:model1}).
\item {\bf Compute relative vectors $\Delta {\bf J}_{(curr,g)}$:} This process requires the continuous prediction of the relative vector $\Delta {\bf J}_{(curr,g)}$ to provide feedback to the user. To achieve this, it is essential to know the current catheter imaging state ${\bf S}_{curr}^{home}$ (Equation\,\eqref{eq:model2}) to calculate the relative vectors needed for movement. Then, using $\mathcal{IK}$, the relative joint vectors $\Delta {\bf J}_{(curr,g)}$ from the current state to the goal state are computed at Equation\,\eqref{eq:model3}. Figure\,\ref{fig:framework}~(b) shows the iterative process.
\item {\bf Iteration:} Repeat the above steps 3-4 until the goal view is successfully reached. Once the goal view is reached, the clinician can repeat the process starting from step 2 to transition from the current view to any other desired view.
\end{enumerate}

 \begin{figure*}[t!]
  \center
	\includegraphics[width=0.9\textwidth]{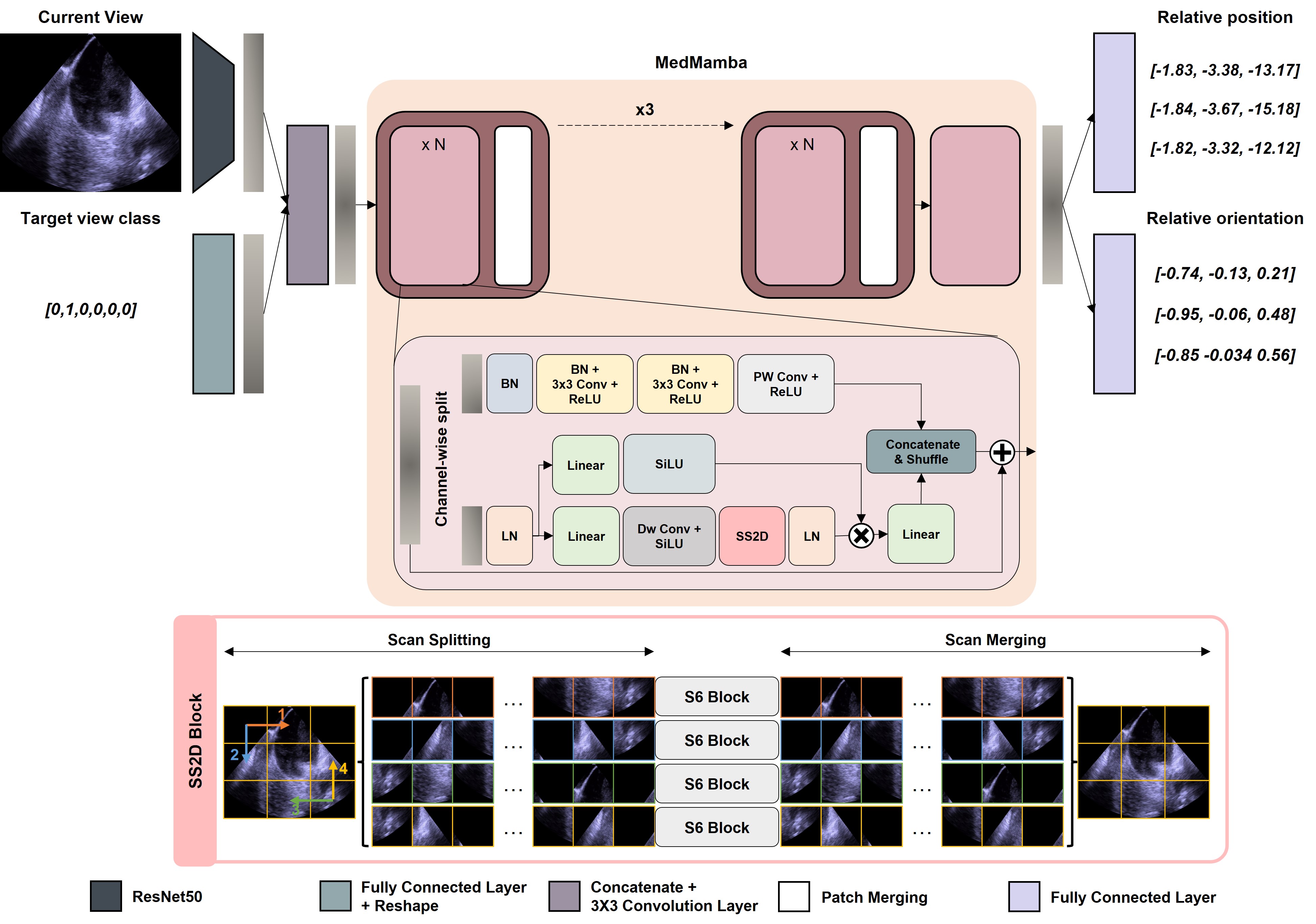}
	\caption{The proposed method architecture. The center of the block represents the MedMamba structure. We have added layers for feature mixing with target class code and image. At the output of the network, we added multi-head structure to estimate position and orientation separately. The details of SS2D structure are shown at the bottom. The image feature is split into sequences in different orders and put into the SSM model. Then, the outputs are merged together.}
 	\vspace*{-0.3cm}
	\label{fig:model}
\end{figure*}

\subsection{Dataset} 
\label{sec:dataset}
To achieve our goal, we need to model a transition function $\mathcal{M}$. For this, we used a dataset collected from a well-established clinical environment and processed using the CARTO mapping system (Biosense Webster Inc., USA)\,\cite{jnj2023carto}. \add{The CARTO system is an application designed for real-time visualization and mapping of cardiac anatomy. It uses an ICE catheter to scan the entire cardiac structure and reconstruct it into a 3-D volumetric mesh using volume contouring algorithms \cite{liao2018more}. It targets the LA, LAA, Left Inferior Pulmonary Vein (LIPV), Left Superior Pulmonary Vein (LSPV), Right Inferior Pulmonary Vein (RIPV), Right Superior Pulmonary Vein (RSPV), and ESO, as shown in Figure\,\ref{fig:dataset}~(a). Notably, it captures cardiac images at the end of the T-wave, providing consistent volumetric information. The CARTO images are labeled according to the anatomical structures they visualize. These images are paired with the corresponding tip position and orientation, which are obtained from the electromagnetic (EM) sensor. The EM sensor provides precise tracking of the catheter's spatial configuration. Consequently, the CARTO system integrates four key elements: US images, anatomical labels, position/orientation information, and the subject's cardiac mesh—all of which were utilized in this research. We obtained the CARTO dataset from a total of 858 subjects. Of these, 793 subjects were allocated for training, 25 for validation, and 40 for testing. This resulted in approximately 143,400 cases for training, 4,870 cases for validation, and 6,532 cases for testing.}

\subsection{Reconstruction of Missing Volume Meshes}
Our dataset mainly consisted of LA, LAA, LIPV, LSPV, RIPV, and RSPV volumes. Thus, it contains only a limited number of samples for the RA, RV, and LV volumes, comprising five subjects. However, the RV and LV are crucial for intra-cardiac examinations, with the RV view often serving as the {\it home view}-the starting point for most procedures.

\add{To address this limitation, we estimated the centers of the RA, RV, and LV volumes based on their relationships with other structures. Specifically, we determined the spatial relationships between each of the RA, RV or LV and the LA and LAA. By extracting these relationships from a subset of the dataset containing all these volumes, we applied the derived relationships to the entire dataset as shown in Figure\,\ref{fig:dataset} (b).}

\subsection{Data pre-processing}
In the dataset, each subject has its own coordinate system, referred to as the world coordinate system. This can vary due to the EM sensors used during data acquisition, leading to inconsistencies across subjects. To address this, we normalized all datasets based on the transducers used in each image, eliminating dependence on the world coordinate system. This normalization is also more practical, as transducer manipulation should be performed relative to the transducer itself.

The transducer state in the world coordinate system can be represented as ${\bf S}_{w}^{tr}$, and its corresponding transformation matrix is $T_{w}^{tr}$. We can obtain the target position and orientation relative to the transducer following equation:
\begin{equation}
    {\bf T}_{tr}^{tar} = ({\bf T}_{w}^{tr})^{-1} \cdot {\bf T}_{w}^{tar},
\end{equation}
where the ${\bf T}_{w}^{tar}$ (${\bf S}_{w}^{tar}$) is target position and orientation, which visualizing each center of the mesh. According to clinical guidance in ICE imaging procedures, the structures are visualized by rotating the ICE catheter with minimal variation in position. Therefore, we selected a single point inside the RA volume for each target view as the target position. To determine the orientation, we aligned the fan direction (from the target position to the center point of the fan's long axis) to point toward the center of the mesh. The final positions and orientations, ${\bf S}_{tr}^{tar}$, were used in our target dataset. The composition of the entire dataset for a single subject is shown in Figure\,\ref{fig:dataset} (c), while the individual targets for a single subject and view are shown in Figure\,\ref{fig:dataset} (d).

\subsection{Network details}

We utilized MedMamba as the base model for our method \cite{yue2024medmamba}. MedMamba combines CNN with Mamba, which uses SSM. This combination leverages both local and global information, showing superior performance in medical classification tasks compared to other conventional methods. Consequently, we applied MedMamba to our view navigation research to solve the regression problem.

The entire MedMamba architecture is shown in Figure\,\ref{fig:model}. It consists of four sequences of SS-Conv-SSM blocks and patch merging steps, except for the last block, which contains only an SS-Conv-SSM block. The SS-Conv-SSM block splits its feature map into two branches: one for local feature extraction using convolution and the other for global feature extraction using the core element, 2D-Selective-Scan (SS2D). It employs the Cross-Scan Module (CSM), which utilizes a four-way scanning strategy to achieve a global receptive field without increasing the computational burden. The selective scan state-space model (S6) captures long-sequence dependencies from all directional features and merges them to construct the 2D feature map. The patch merging step reduces the spatial dimension to achieve hierarchical representations.
 
To achieve our goal, we modified both the network architecture and the training loss. The original MedMamba was designed for medical image classification, which required only a single image as input. However, our objective was to develop a multi-target regression model that not only takes an image as input but also incorporates the target class. To facilitate this, we introduced a feature-mixing layer, consisting of a single convolutional layer. Specifically, the ICE image, representing the current view, is fed into the network and processed through ResNet50 to extract features \cite{he2016deep}. Simultaneously, the target view class, encoded as a one-hot vector of size 6 (corresponding to the number of target views), is passed through a linear layer that adjusts its dimensions to match the squared feature size of the ResNet50 output. This vector is then resized and concatenated with the image features before being passed through the feature-mixing layer. Additionally, to estimate both position and orientation effectively, we added two linear layer heads at the output of MedMamba: one for position regression and the other for orientation regression.

The entire network was trained using quantile loss to account for uncertainty and variability, enabling it to effectively estimate the boundaries of predictions. This approach allows the network to estimate not only the median value but also the $q^{th}$ percentile, providing the boundary of estimation with $q^{th}$ reliability. The simplified loss function is shown below:
\begin{equation}
l_{quant}(y, \hat{y}) = \left \{\begin{array}{ll}
\alpha \cdot (y-\hat{y}), & \hat{y} \leq y \\
(1-\alpha) \cdot (\hat{y}-y), & \hat{y} > y,
\end{array}
\right.
\end{equation}
where the $\alpha$ denote the percentage of $q^{th}$ quantile, y and $\hat{y}$ represent elements of label vector $\mathbf{y}$  and prediction vector $\mathbf{\hat{y}}$, respectively. 
The total loss function used to train the network is shown below:
\begin{equation}
l_{total} = l_{quant}(p, \hat{p}) + \lambda * l_{quant}(o, \hat{o}),
\label{eq:total_loss}
\end{equation}
where $p$ and $o$ represent the labels for position and orientation, respectively, and $\hat{p}$ and $\hat{o}$ denote the predicted position and orientation, respectively. The parameter $\lambda$ is the weighting factor.

\section{System Details and Evaluation Metrics}
\label{sec:implementation}

 \begin{figure}[t!]
  \center
	\includegraphics[width=0.47\textwidth]{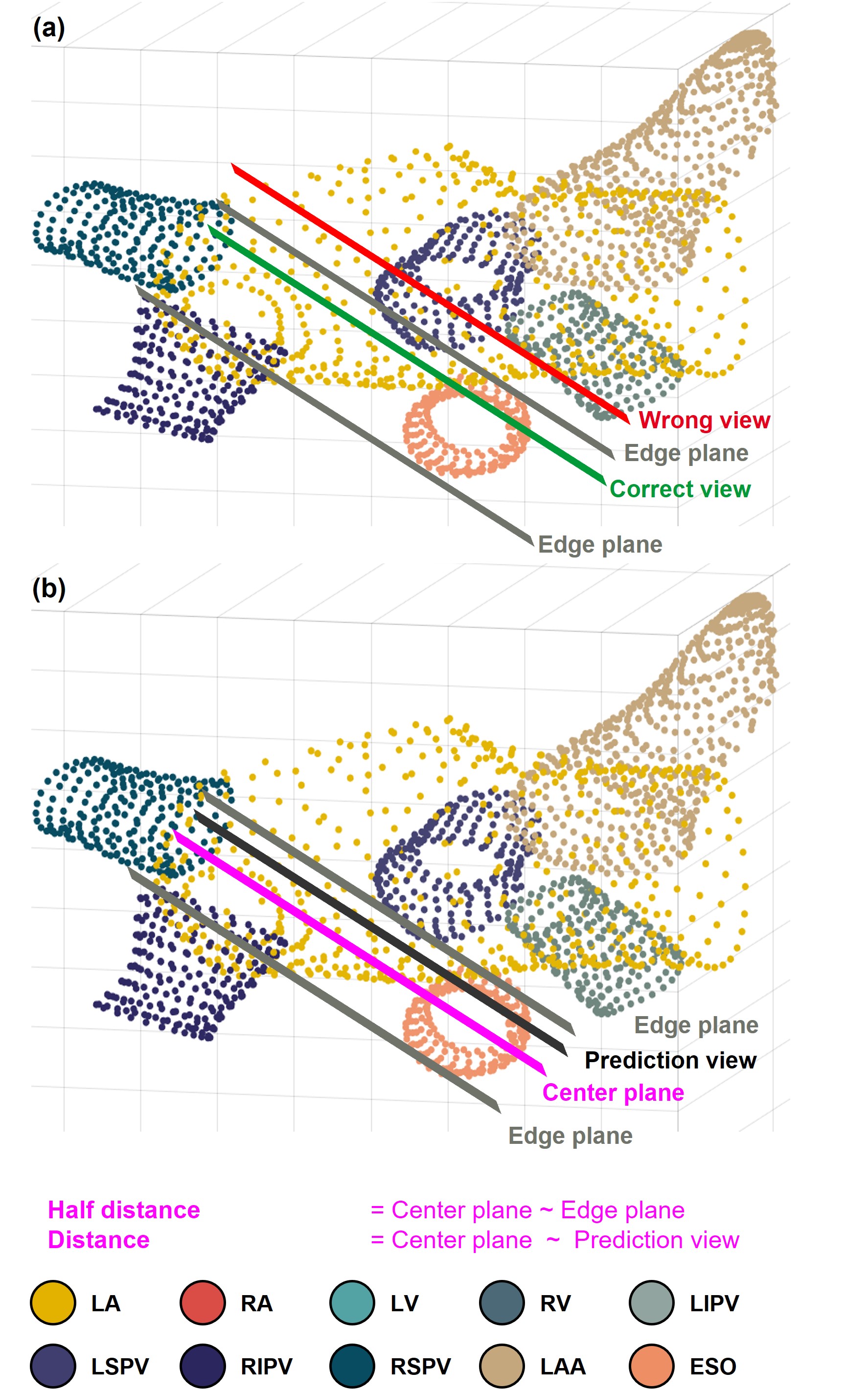}
	\vspace*{-0.1cm}
	\caption{Validation method. (a) \rev{Percentage of correct views} calculation: The prediction is considered correct if the predicted view matches the target volume, and incorrect otherwise. \add{(b) Distance calculation: The distance is measured from the center of the target volume to the predicted view.}}
 	\vspace*{-0.3cm}
	\label{fig:validation}
\end{figure}

 \begin{figure*}[t!]
  \center
	\includegraphics[width=\textwidth]{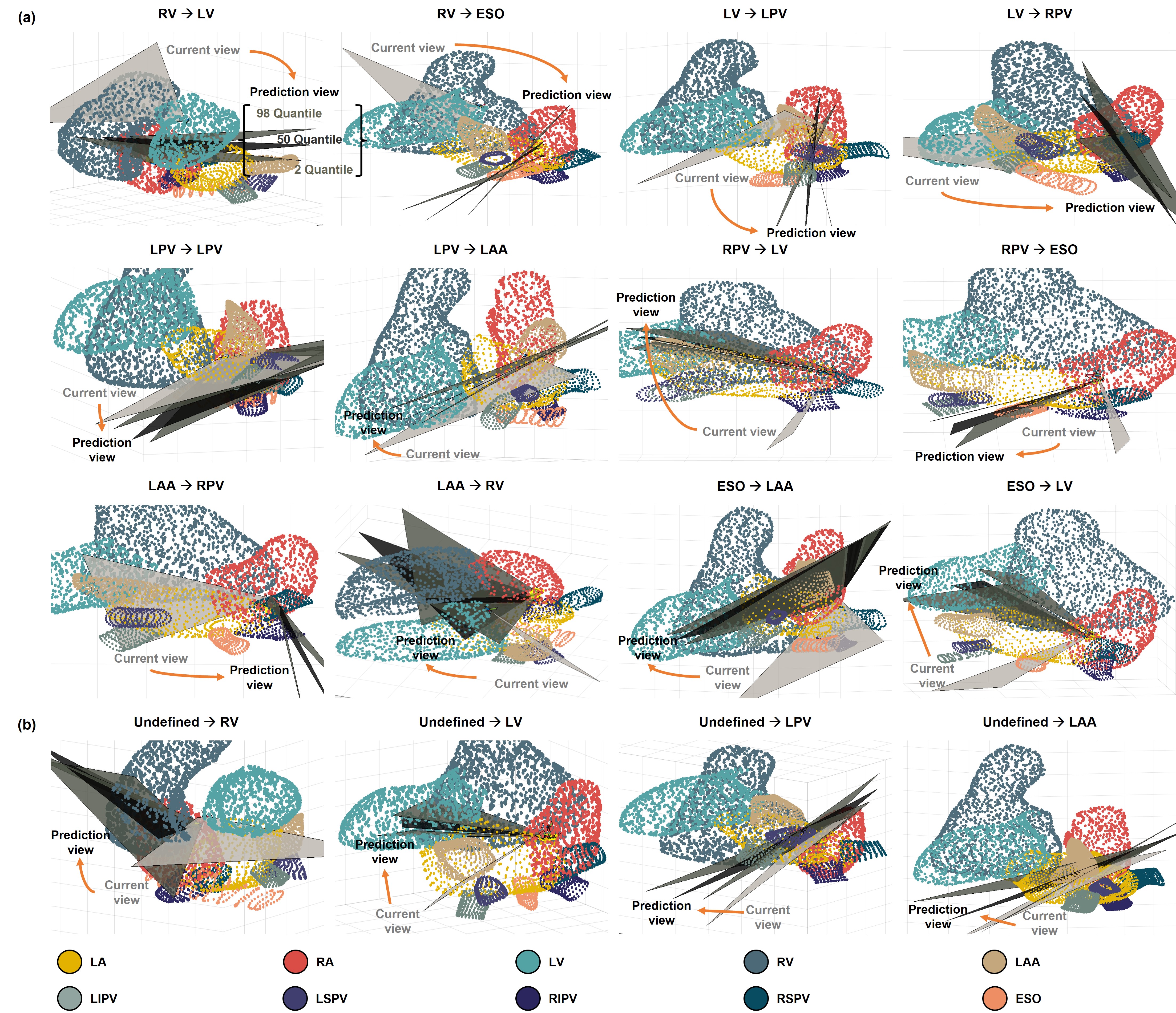}
	\vspace*{0.2cm}
	\caption{Random view to clinically defined view results. The light gray is current view, and the black denotes the $50^{th}$ quantile prediction view. Each of dark gray denote the $98^{th}$, $2^{nd}$ quantile prediction views, representing the boundaries of the prediction. The orange arrow indicates the transition from the current view to the predicted view, based on the output of the proposed network. (a) \rev{Results using 12 randomly selected samples from the anatomically labeled test set, each visualizing a specific structure.} (b) \rev{Results using 4 randomly selected samples from the test set labeled as undefined views, each visualizing non-specific structures only crossing the LA.}}
	\label{fig:qualitative}
\end{figure*}

\subsection{Evaluation metric}
\label{sec:evaluation}
To evaluate the proposed method, we aimed to determine how accurately the relative position and orientation could guide the fan to the target mesh. We assessed this in two ways. First, we used classification to measure how well the fan's position and orientation, as predicted by the network, align with the target volume. As shown in Figure\,\ref{fig:validation} (a), a correct view is defined as the fan being properly positioned within the target volume, while a view where the fan is outside the target volume is considered incorrect. We then calculated the \rev{percentage of correct view across all test cases.}

\add{Secondly, we measured the distance between the proposed fan position and the center of the target volume. As shown in Figure\,\ref{fig:validation} (b), we calculated the vertical distance between the center of the target volume and the predicted fan position.}


\subsection{Training details}
The network was trained using the Adam optimizer with a learning rate of 1e-4 for 580 epochs. We employed a batch size of 40, and the entire image dataset was resized to $224 \times 224$. In Figure\,\ref{fig:model}, $N$ is set to [2,2,4,2] for each block, which is the default setting of MedMamba-Tiny. The $\lambda$ in Equation\,\ref{eq:total_loss} is 10, and $\alpha$ is 0.98, which denotes the $98^{th}$ quantile. The entire framework was implemented in PyTorch and trained on a single A100 GPU.

\subsection{{Guidance system frame rate}}
\add{The proposed method provides continuous guidance within a closed-loop system, ensuring timely updates for effective navigation. To evaluate its performance in this regard, we measured the inference time using a single A100 GPU, achieving a frequency of 50 \rev{[fps]}.}

 \begin{figure*}[t!]
  \center
	\includegraphics[width=0.97\textwidth]{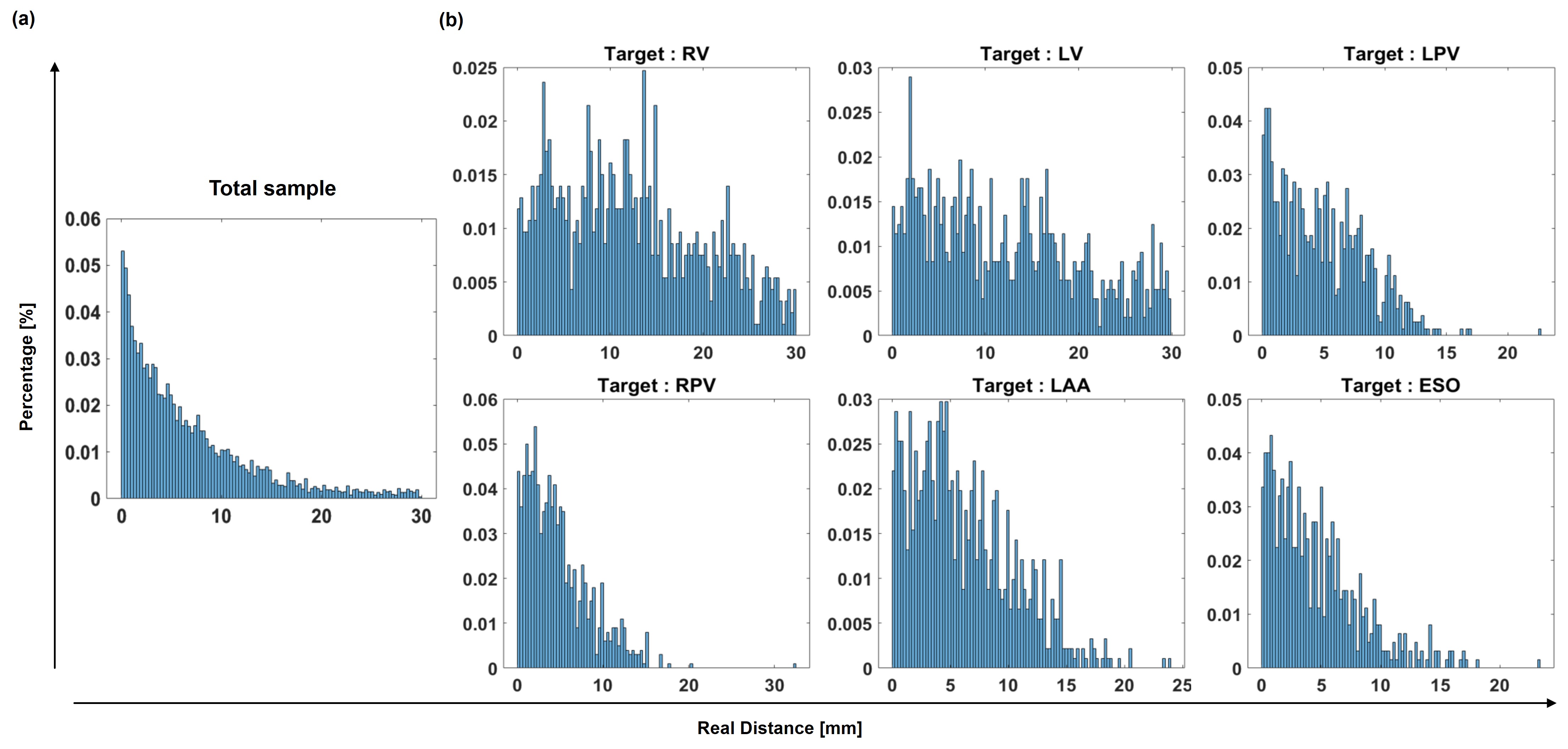}
	\vspace*{-0.3cm}
	\caption{\add{Histogram of the distance between the center of the target mesh and the predicted fan. The x-axis represents the distance, and the y-axis represents the percentage of samples. (a) Real distance distribution for all samples. (b) Real distance distribution for each target view.}}
	\label{fig:quantitative}
\end{figure*}

\section{Experimental Results and Analysis}
Since the proposed method is specifically designed for real clinical applications and trained on actual clinical data, direct validation in real-world scenarios is limited. Therefore, we primarily present simulation-based results derived from the real clinical CARTO dataset, evaluated both qualitatively and quantitatively. These simulation-based evaluations assess the method's performance by comparing predicted values with the corresponding volume mesh. For example, we adjusted the fan’s position and orientation based on the predicted values and verified their alignment with the associated volume mesh. Several simulation experiments were conducted, as shown below:
\begin{enumerate}[label=\Alph*]
    \item Random view to clinically-defined view: This experiment aims to verify whether the position and orientation provided by the proposed method effectively guide the user to the desired target view.
    \item Distance error between prediction and ground truth: This experiment aims to verify whether the model has been well-trained on the training dataset by evaluating the accuracy of the predicted position and orientation compared to the ground truth.
    \item Distance error between the center of the target mesh and the predicted fan: This experiment evaluates whether the predicted fan accurately intersects the target mesh by measuring its proximity to the center of the mesh.
    \item Distance error from a nearby view to a clinically-defined view: This experiment assesses whether the proposed method provides consistent and similar prediction values when the current views are located nearby.
    \item Distance error for intermediate views while moving towards the target view: This experiment verifies whether the proposed method can effectively guide the user to the target mesh in dynamic scenarios.
\end{enumerate}
To further validate the effectiveness of the proposed method in a realistic scenario, we conducted a semi-simulation experiment as described below:

\begin{enumerate}[label=\Alph*]
  \setcounter{enumi}{5}
  \item Human-in-the-loop validation: This experiment assesses whether the proposed method can be effectively applied in realistic scenarios, allowing engineers to directly manipulate the ICE catheter based on the guidance provided.
\end{enumerate}

The entire validation procedure was simulated using MATLAB 2022a, including visualization and validation of the sensor data with the CARTO dataset. Specific details of the simulation environment are provided in each section.

We evaluated our method using 40 subjects, none of whom were part of the training dataset. Each subject included several frames, resulting in a total of 6,532 test cases used in the final test set.



\subsection{Random view to clinically-defined view}
\rev{The dataset we used includes anatomical labels corresponding to each ultrasound image. Among the 6,532 test cases, there are several cases labeled as RV, LV, LAA, LPV, RPV, ESO, and undefined views, with the undefined view only crossing the LA and not any other structure. We randomly selected 12 cases—two for each view—and 4 cases for the undefined view.}

For each selected case, the proposed method provided the relative position and orientation required to achieve the target view. Based on these predicted states, we adjusted the fan's current position and orientation to validate whether the predicted fan accurately crossed the target mesh.

The qualitative results are shown in Figure\,\ref{fig:qualitative}. The light gray color indicates the current view fan, including position and orientation information, while black and dark gray represent the $50^{th}$ and $98^{th}$ or $2^{nd}$ percentiles of the predicted fan, respectively.

In all cases, the predicted fan is clearly contained within the target volume. Notably, for cases requiring significant movements, such as transitioning from the LV to the RPV, the predictions accurately align with the desired target state. Similarly, for cases involving minimal movement, such as navigating from the LPV to the LPV, the predictions effectively support precise view navigation. In Figure\,\ref{fig:qualitative} (b), which demonstrates the results of transitions from undefined views to clinically-defined views, the proposed method successfully provides accurate target state information for both small and large movements.

\subsection{Distance error between prediction and ground truth}
We calculated the errors between the predicted values, such as position and orientation, and their corresponding ground truth. Predictions were obtained for all 6,532 test cases, and the errors were computed accordingly. The average position error was 9.75 [mm], with a standard deviation of 8.50 [mm]. For orientation, the average error was (3.48, 1.13, -1.87) [deg], with a standard deviation of (47.27, 22.24, 36.51) [deg]. These results demonstrate that the proposed network has been trained effectively, as it accurately predicts the target position and orientation based on the current view image. The relatively high average orientation error is attributed to cases where the target and current positions are close, such as the RV and LV. In such cases, the target's large size allows for a wider range of acceptable orientations. This results in a high standard deviation in orientation; however, as demonstrated in Section\,\ref{sec:accuracy}, the method maintains high \rev{percentage of correct views} despite this variability.

\subsection{Distance error between the center of the target mesh and predicted fan}
\label{sec:accuracy}
We used the evaluation method described in Section\,\ref{sec:evaluation}. Based on the predicted position and orientation, we adjusted the current view fan to align with the predicted fan. The proposed method also provides prediction boundaries, and we evaluated \rev{the percentage of corrects views} by assessing how well the predicted quantile planes were contained within the target volume. Furthermore, we measured the distance between the predicted $50^{th}$ quantile planes and the center of the target mesh, with smaller distances indicating better predictions.

\rev{The proposed method achieved an 89\%, demonstrating its strong performance in effectively guiding views.}

Figure\,\ref{fig:quantitative} (a) shows the real distance histogram for the entire test dataset, including out-of-boundary (OOB) samples. In Figure\,\ref{fig:quantitative} (b), it gives the correct sample distance distribution for each target view, visualized as histograms for various input images corresponding to target views such as RV, LV, LPV, RPV, LAA, or ESO. For the RV and LV target views, the histograms are slightly noisier due to the smaller number of training samples compared to other target views. However, consistent with the analysis that smaller distance correlate with more accurate predictions, this pattern is evident in all cases.

  \begin{figure*}[t!]
  \center
	\includegraphics[width=\textwidth]{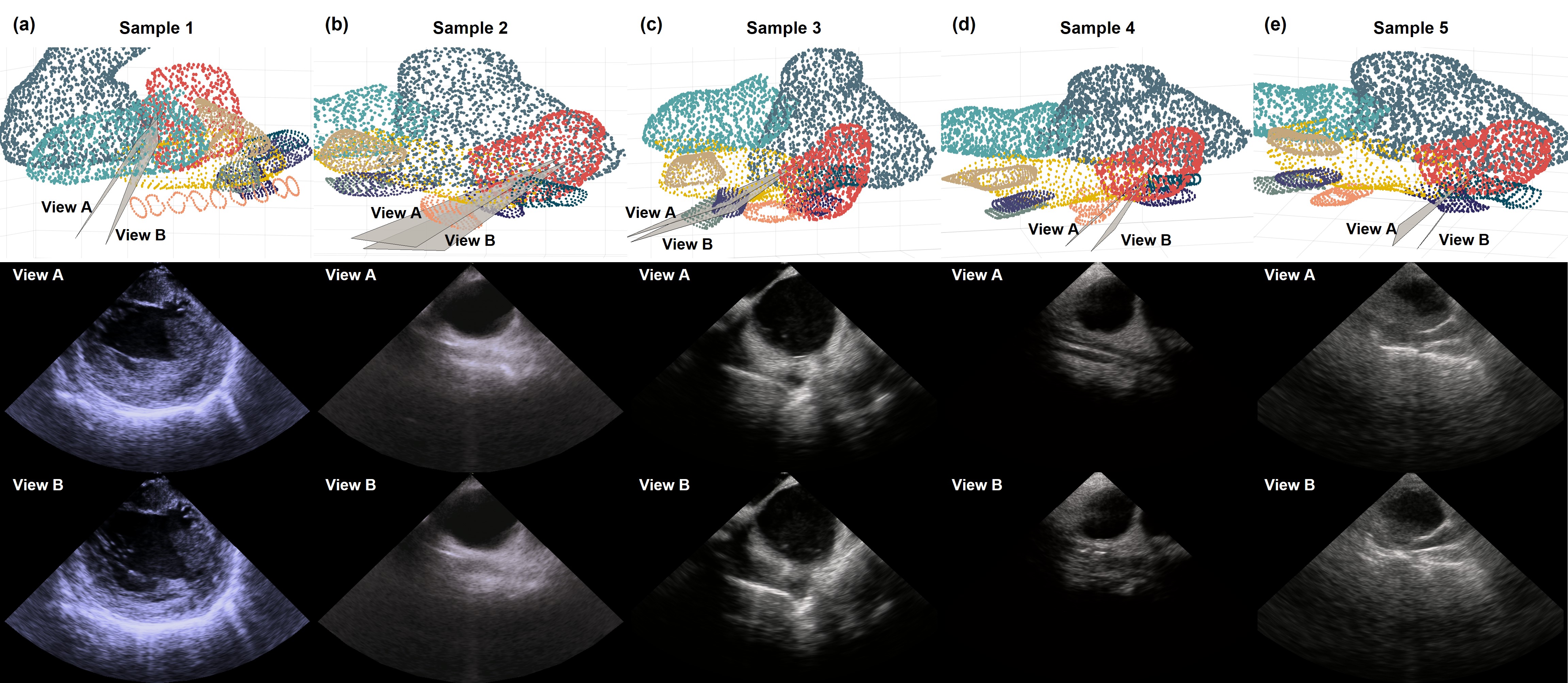}
	\caption{Distance error from a nearby view to clinically-defined view. 'View A' and 'View B' show similar or nearby view of the same area. (a) Both images show the LV view (b) Both images show the ESO view (c) Both images show the LPV view. (d) 'View A' shows an undefined view, and 'View B' shows the ESO view. (e) Both images show the RPV view.}
 	\vspace*{0.3cm}
	\label{fig:nearby}
\end{figure*}

\begin{table*}[h!]
    \centering
    \caption{Distance error from a nearby view to clinically-defined view}
    \vspace{0.3cm}
	\resizebox{\textwidth}{!}{
		\begin{tabular}{c||c|c|c|c|c|c}
			\hline 
			     & RV & LV & LPV & RPV & LAA & ESO \\ \hline\hline
			Sample 1& 5.72[mm] & 6.18 [mm]  & 7.73 [mm] & 10.65 [mm]  & 8.03[mm] & 9.95[mm] \\ 
            (LV view)  & (0.045,0.121,0.088)[rad]  & (0.079,0.138,0.105)[rad]  &(0.079,0.085,0.054)[rad]  & (0.073,0.064,0.154)[rad]  & (0.028,0.077,0.090)[rad]  & (0.075,0.089,0.145)[rad]  \\   \hline
            Sample 2 & 3.83[mm] & 3.24 [mm]  & 2.80 [mm] & 3.55 [mm] & 2.86[mm] & 3.33[mm] \\ 
            (ESO view) & (0.038,0.176,0.021)[rad]  &  (0.089,0.065,0.025)[rad]  &  (0.001,0.027,0.072)[rad]  & (0.082,0.045,0.095)[rad]  &  (0.029,0.013,0.150)[rad]  & (0.061,0.028,0.051)[rad]  \\  \hline
            Sample 3 & 4.12. [mm]  & 3.92 [mm] & 3.59 [mm] & 2.96 [mm] & 2.92 [mm]  & 3.19 [mm]  \\ 
            (LPV view) & (0.289,0.088,0.234)[rad]  & (0.095,0.052,0.059)[rad]  & (0.025,0.079,0.124)[rad]  & (0.114,0.192,0.050)[rad]  & (0.073,0.187,0.149)[rad]  & (0.013,0.067,0.054)[rad]  \\  \hline
            Sample 4 & 3.91 [mm] & 1.94 [mm]  & 2.83 [mm] & 3.39 [mm] & 3.53 [mm]  & 2.05 [mm] \\ 
            (Undefined \& ESO view) & (0.169,0.103,0.466)[rad]  & (0.089,0.108,0.095)[rad]  & (0.056,0.002,0.084)[rad]  & (0.000,0.041,0.098)[rad]  & (0.040,0.039,0.086)[rad]  & (0.097,0.017,0.060)[rad]  \\  \hline
            Sample 5 & 3.34 [mm] & 3.07 [mm]  & 3.34 [mm] & 2.67 [mm] & 3.94 [mm]  & 3.37 [mm]  \\  
            (RPV view) & (0.129,0.010,0.052)[rad]  & (0.101,0.008,0.017)[rad]  & (0.173,0.039,0.002)[rad]  & (0.088,0.022,0.092)[rad]  & (0.155,0.003,0.011)[rad]  & (0.144,0.008,0.005)[rad]  \\  \hline
		\end{tabular}
	}
    \vspace{0.5cm}
    \begin{tablenotes}
        \item The upper value represents the difference in predicted position between the first and second views, while the lower value represents the difference in orientation between them. Each value in parentheses indicates the difference along the x-, y-, and z-axes.
    \end{tablenotes}
    \vspace{0.5cm}
	\label{table:comparison}
\end{table*}

\subsection{Distance error from a nearby view to clinically-defined view}

We randomly selected two views from five samples in different subjects, each visualizing similar or nearby areas. For each sample, we selected a target view and obtained the prediction values for each nearby view. We then calculated the error between the prediction values, with the results shown in Figure\,\ref{fig:nearby} and Table\,\ref{table:comparison}.

Figure\,\ref{fig:nearby} shows examples of nearby or similar views. In each sample, 'View A' and 'View B' represent images of very similar areas but captured from different perspectives. Using our proposed method, we obtained the predicted target states for six target views from both images and computed the differences between them. The summarized results are presented in Table\,\ref{table:comparison}.

In the case of Sample 1, both views visualize the LV. When targeting the RV viewpoint, the difference between the outputs of the two views is $5.72[mm]$ and $(0.045,0.121, 0.088)[rad]$, demonstrating a very similar result. In most cases, the results show a good match in position and orientation differences for each target view. These findings demonstrate that the proposed method effectively accounts for the difference in images and translates it into differences in coordinates.

\subsection{Distance error for intermediate view while moving to the target view}
When a user moves the ICE catheter from the current view to the target view based on the proposed method's guidance, the system should continuously guide the user through all intermediate view images toward the target. To validate this, we tested our method using a sequence of images representing the movement from the starting point to the target point. The predicted fan direction was visualized at each step during the transition. For this evaluation, we selected four random subjects and utilized a series of images, each representing states ranging from an arbitrary starting point to a specific target volume.
 
\begin{figure*}[t!]
  \center
	\includegraphics[width=0.98\textwidth]{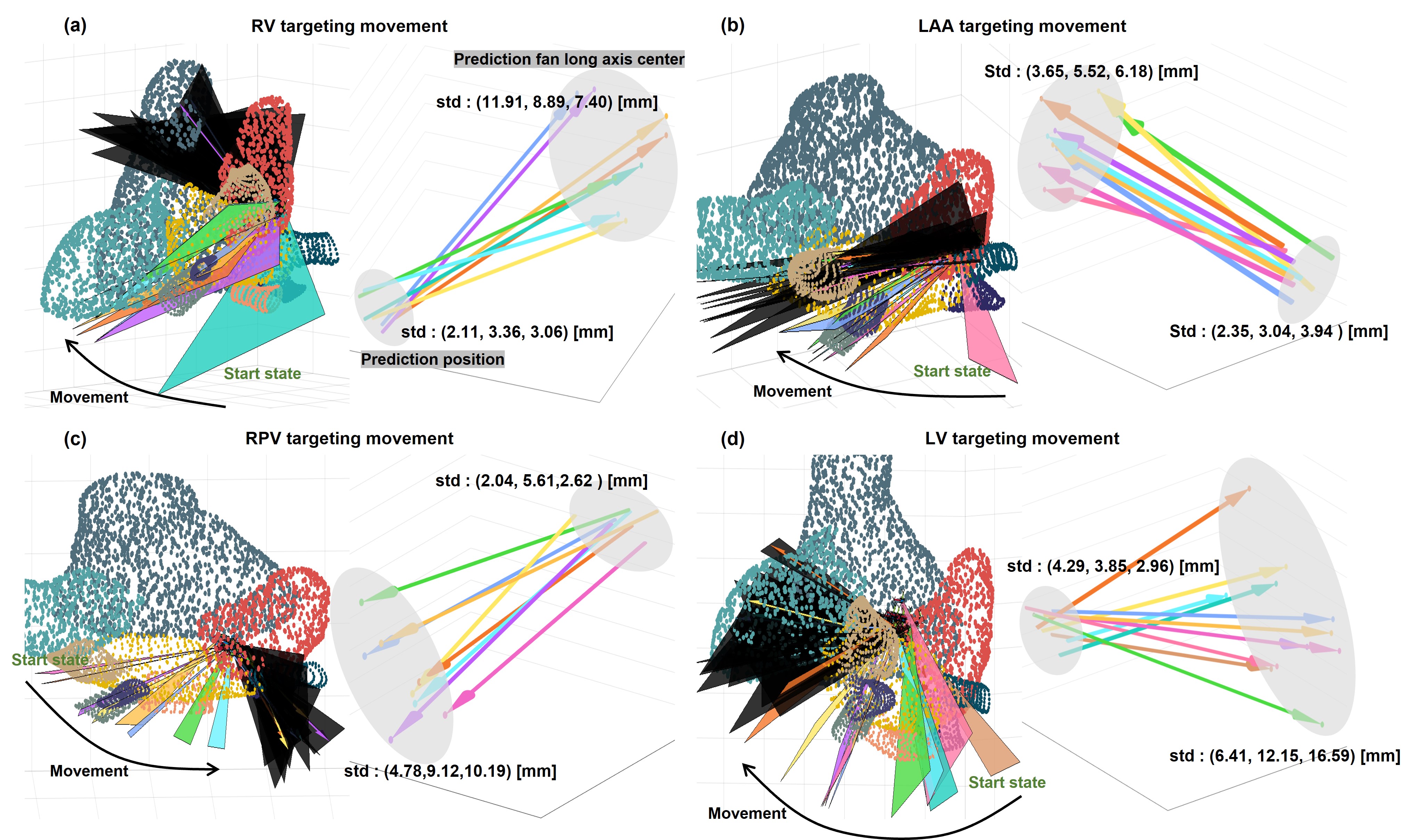}
	\caption{Distance error for intermediate views while moving to the target view. For several intermediate images during the movement from the starting point to the target view, we acquired the target state from the proposed method and determined its fan direction. The left side shows the fan movement and its predicted position, while the right side shows the predicted fan direction vector. (a) RV target movement, (b) LAA target movement, (c) RPV target movement, and (d) LV target movement.}
	\label{fig:state}
\end{figure*}

The results are shown in Figure\,\ref{fig:state}. The colored triangles indicate the fan of the current image, while the black triangles represent the predicted fan. We determined the fan direction by measuring from the predicted position to the center point of the fan's bottom long axis. This fan direction is separately visualized on the right side of each volume and fan image. Each standard deviation of the predicted position and the center point of the fan's bottom long axis is calculated and displayed next to the region, highlighted by a light gray elliptical shape.

In Figure\,\ref{fig:state} (a), we assumed movement from the blue fan point to the RV target. From all intermediate images, we acquired the predicted fan direction, visualized as a black fan. The complete fan direction is shown on the right side, showing that all predicted fan directions from intermediate images are consistently aligned. In Figure\,\ref{fig:state} (c), we assumed movement from the pink fan point to the RPV target. The predicted fan directions from all intermediate images are shown on the right side. It demonstrates that the proposed method provides a stable target state for all intermediate images during movement and effectively functions in a closed-loop system.

\subsection{Human-in-the-loop validation}

To validate our method in a realistic scenario, we conducted a discrete action guidance-based human-in-the-loop experiment. Since the method is designed and trained using real cardiac data, direct testing on human subjects was not feasible due to the inability to acquire real clinical imaging at arbitrary positions and orientations.

\begin{figure}[t!]
  \center
	\includegraphics[width=0.5\textwidth]{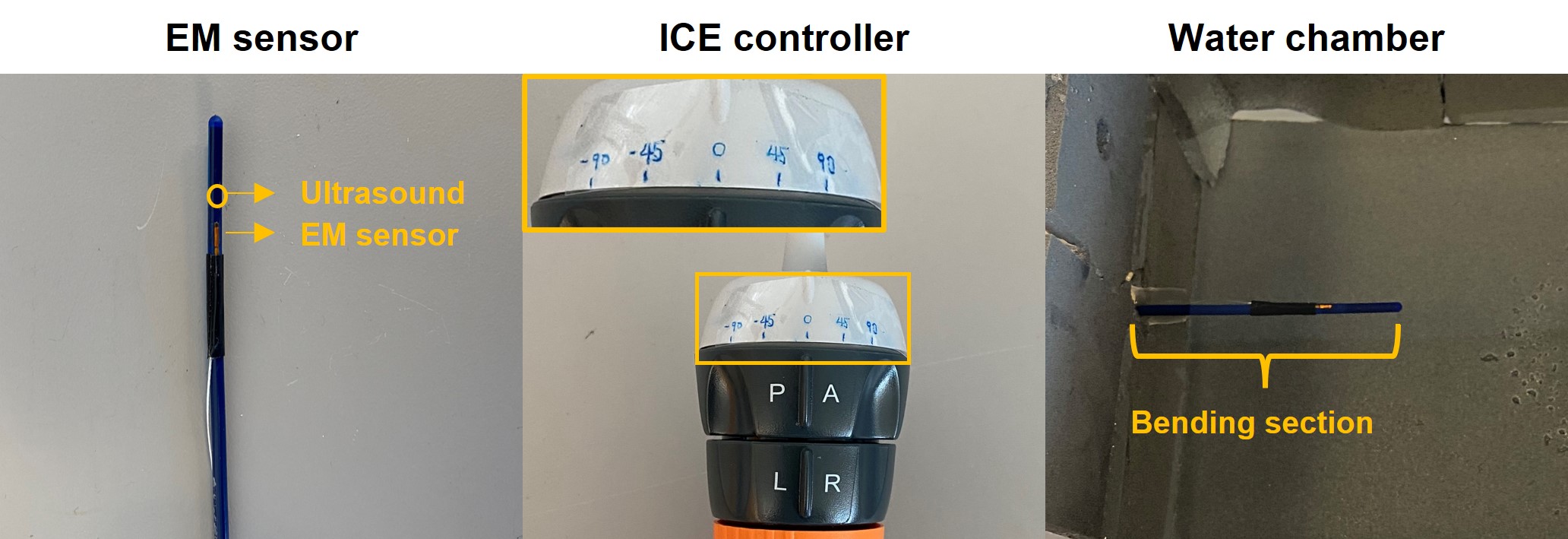}
	\vspace*{0.1cm}
	\caption{Human-in-the-loop validation setup: The EM sensor was attached to the ICE catheter at the transducer location. The ICE catheter was controlled using the ICE controller, which includes knobs for adjusting the AP and LR directions, as well as for rotation and translation. For the experiment, the ICE catheter was positioned in a water chamber, as shown in the last column.}
	\label{fig:human}
\end{figure}

To address this limitation, we implemented a semi-simulation experiment using a real ICE catheter equipped with an EM sensor, as shown in Figure\,\ref{fig:human}. The EM sensor served two purposes: 1) identifying the nearest real cardiac image from the CARTO dataset based on the ICE catheter's current position and orientation, and 2) calculating errors relative to the target pose for performance validation. By leveraging the CARTO dataset, which contains pre-recorded clinical imaging, the system was able to map real clinical images to the ICE catheter's pose at discrete intervals, simulating a realistic surgical environment. This approach, termed ``semi-simulation'', combines real catheter manipulation with CARTO-derived imaging guidance. While the catheter is controlled in a physical setting, clinical imaging is mapped at discrete steps using CARTO data, compensating for the absence of direct clinical imaging acquisition. This hybrid approach bridges the gap between fully simulated setups and real-world clinical scenarios, providing a practical validation framework for the proposed system.

To further support the user’s adjustments, we incorporated angle markings on the ICE catheter to indicate the degree of knob rotation and bending. These markings allowed users to visually confirm the catheter’s movement during manipulation (shown in Figure\,\ref{fig:human}). Although such angle-specific markings and mechanisms are not typically present in clinical environments, they were introduced as assumptions to establish a clear connection between the user’s manual adjustments and the proposed 4-DoF algorithm. This approach mimicked a real surgical scenario while leveraging CARTO data to provide realistic imaging guidance.

In more detail, the ICE catheter is initially placed in a straight condition as the home view ${\bf S}_{home}$. A target view (${\bf S}_{goal}$) is then randomly selected, and the proposed method provides guidance (${\bf J}_{home}^{goal}$) to assist the test engineers in manually manipulating the ICE catheter. The test engineers adjust the catheter based on the guidance provided on the display panel. When they believe the catheter has aligned with ${\bf J}_{home}^{goal}$, they confirm the position to the proposed system. If the catheter successfully reaches the target view (${\bf S}_{goal}$), the experiment concludes. Otherwise, if the moved state (${\bf S}_{curr}$) does not align with the target view, the proposed method iteratively generates new guidance (${\bf J}_{curr}^{goal}$) to direct further adjustments from the current view to the target view. This process is known as the re-planning step. All decisions and instructions are determined and provided by the proposed method. The procedure was performed on 10 cases from 10 different subjects, with three test engineers participating. The detailed steps of the procedure are shown below:

\begin{enumerate}
    \item Place the ICE catheter in the straight condition as the home view ${\bf S}_{home}$ at a water chamber, and select the RV image of the randomly chosen subject from the real cardiac images in the CARTO testing dataset.
    \item Choose randomly the target view (${\bf S}_{goal}$) from the 6 clinically defined views. 
    \item Proposed method displays (${\bf J}_{home}^{goal}$) on the display panel.
    \item The test engineer manipulates the ICE catheter according to the guidance from the display panel.
    \item The test engineer confirms to the proposed system once they have completed the manipulation of the ICE catheter to (${\bf S}_{goal}$)
    \item If the proposed system confirms (${\bf S}_{curr} \approx {\bf S}_{goal}$), then proceed to step 7. Otherwise, during the re-planning phase, return to step 4 with updated guidance, (${\bf J}_{curr}^{goal}$), based on finding the nearest real cardiac images from the CARTO dataset using position and orientation data. 
    \item Calculate the error between the final state and the CARTO ground truth.
\end{enumerate}

\begin{figure}[t!]
  \center
	\includegraphics[width=0.45\textwidth]{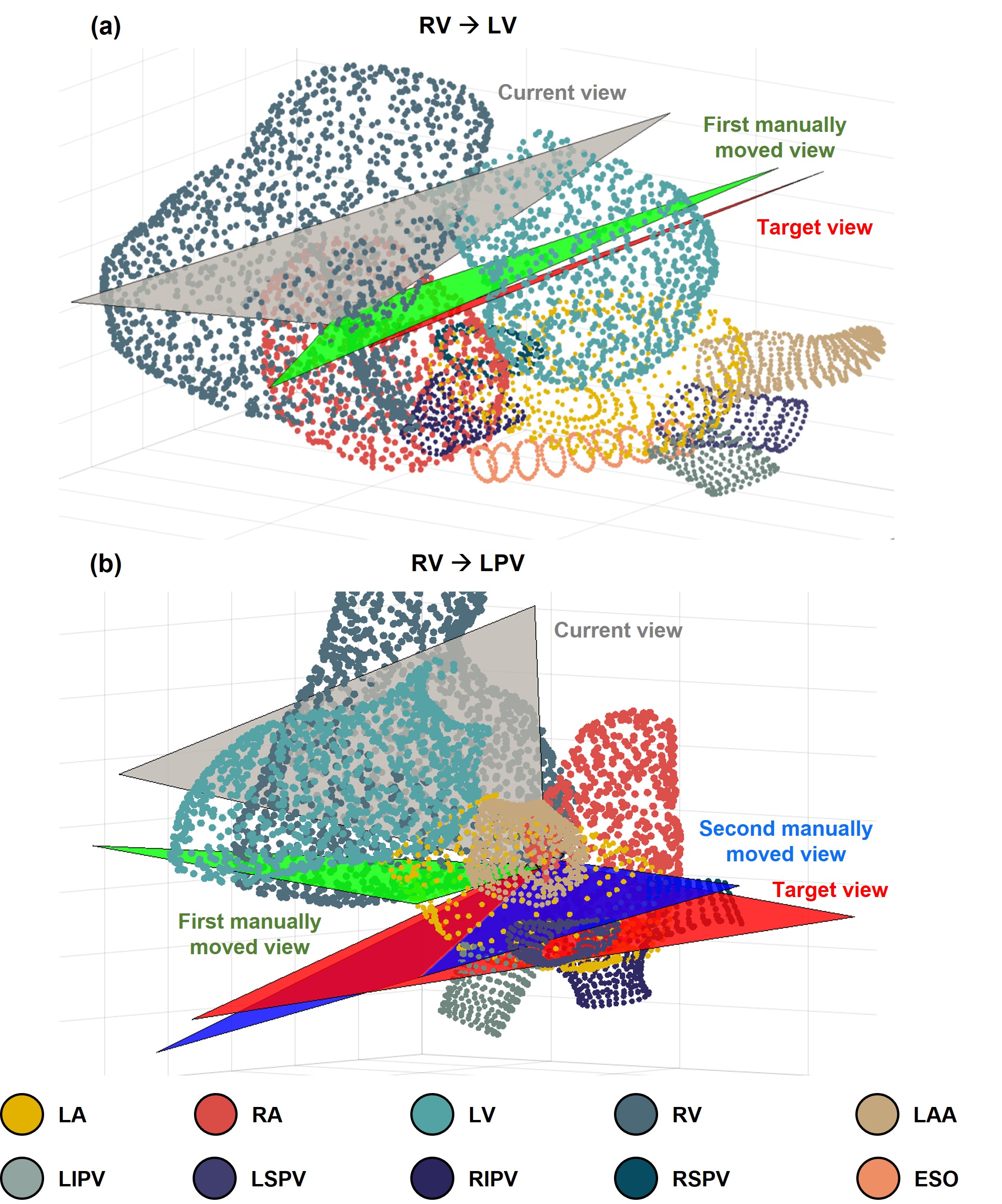}
	\caption{Human-in-the-loop validation results. The gray fan represents the initial view directed toward the RV. The green and blue fans denote the first and second manually adjusted fan views, respectively, while the red fan indicates the target view. (a) Direct movement from RV to LV. (b) Movement from RV to LPV through two re-planning steps.}
	\label{fig:humanloop}
\end{figure}

The representative results are shown in Figure\,\ref{fig:humanloop}. The blue fan represents the manually moved fan, the red fan indicates the target fan, and the gray fan shows the initial view. Figure\,\ref{fig:humanloop} (a) shows a successful result after one movement, where the moved fan successfully crosses the target volume and aligns closely with the target fan. Across all cases, the average error between the final state and the real target value was $6.74[mm]$ and $(10.28, 16.88, 7.25)[deg]$ at step 7, confirming the applicability of our method in a real-world human-in-the-loop system.

Notably, we highlight the re-planning performance, where the error decreased after the re-planning step (Step 4-6). As shown in Figure\,\ref{fig:humanloop} (b), the blue fan, representing the second moved fan, successfully crossed the target volume and aligned closely with the red target fan. The average error for the first movement was $15.12[mm]$ and $(12.02, 17.49, 14.41)[deg]$, which decreased to $4.30[mm]$ and $(6.98, 10.94, 3.78)[deg]$ after re-planning. These results demonstrate that the proposed method can effectively support continuous view navigation guidance through re-planning, ensuring precise movements and preventing missed alignments in a human-in-the-loop system.

\section{Conclusion}
ICE is an imaging tool used to capture images of cardiac structures within the heart chambers. It is valuable for anatomical assessments, diagnosing congenital heart defects, and evaluating valvular heart disease. Additionally, ICE plays a crucial role in interventional cardiology, particularly in guiding catheter ablation and the closure of septal defects. Despite its importance, effective manipulation of the ICE requires a skilled operator and extensive training due to the difficulty in achieving precise and repeatable control. To address this challenge, we propose an AI-driven closed-loop view guidance system designed to simplify ICE manipulation based on the current ICE image. Specifically, there are clinically predefined common views for most procedures, such as RV, LV, LPV, RPV, LAA, and ESO views. Our method provides the relative position and orientation vectors necessary to guide the user to their preferred viewpoint from the current view, facilitating easier control of the ICE.

To achieve this, we developed a Mamba-based regression model. The Mamba model, built upon the SSM, outperforms traditional transformer architectures. Its selective scan and hardware-aware features allow for efficient handling of long-range sequences. Our approach utilizes the MedMamba structure, which combines CNN and Mamba to effectively capture both global and local features in medical images. For multi-target class regression, we incorporated a feature mixing layer to combine image features with target class codes. Moreover, we added a multi-head structure at the network's end to separately estimate position and orientation. The network was trained using quantile loss to accurately estimate the prediction boundaries. We trained our regression model on a CARTO dataset that includes view, position, orientation, and cardiac volume information.

The qualitative and quantitative results demonstrate that our method provides accurate target state information for view visualization. Whether the view focuses on specific structures or undefined regions, our method effectively supports view navigation. In particular, the human-in-the-loop validation emphasizes the feasibility of the proposed method in realistic scenarios. Although a gap between the real CARTO dataset and the EM sensor introduces some error, the overall results confirm the effectiveness and robustness of our method.

The human-in-the-loop validation demonstrates that our method can seamlessly integrate into real surgical procedures. To further enhance human-machine interaction, a speech or gesture recognition module can be incorporated, allowing for more intuitive control. Users can select from one of six representative views displayed on the interface panel, with view status updates provided at \rev{50 [fps]}, matching the inference time of the method. The control panel displays real-time joint state information for four DoF and offers step-by-step guidance for adjusting each DoF. The system operates continuously until the user issues an end command.

For future work, we plan to validate our system's performance using a beating heart phantom. We believe that our method will simplify ICE manipulation for clinicians and enhance the overall workflow of the procedure.

\section*{Disclaimer}

The concepts and information presented in this paper are based on research results that are not commercially available. Future availability cannot be guaranteed.

\bibliography{ref}

\vfill

\end{document}